\documentclass[runningheads]{llncs}
\usepackage[T1]{fontenc}
\usepackage{graphicx,verbatim}
\usepackage{algorithmic}
\usepackage{algorithm}
\usepackage{booktabs}
\usepackage{multirow}
\usepackage{graphicx}
\usepackage{gensymb}
\begin{document}
\title{Towards Robust Surgical Automation via Digital Twin Representations from Foundation Models}
\author{Hao Ding\inst{1} \and Lalithkumar Seenivasan\inst{1} \and Hongchao Shu\inst{1} \and Grayson Byrd\inst{1} \and Han Zhang\inst{1} \and Pu Xiao\inst{1} \and Juan Antonio Barrag\inst{1} \and Russell Taylor\inst{1} \and Peter Kazanzides\inst{1} \and Mathias Unberath\inst{1}}
\authorrunning{H. Ding et al.}
\titlerunning{DT-Automation}
%
\institute{Johns Hopkins University, 3400 N Charles St, Baltimore, 21218, MD, USA \\ \email{\{hding15,unberath\}@jhu.edu}}

\maketitle              
\begin{abstract}
Large language model-based (LLM) agents are emerging as a powerful enabler of robust embodied intelligence due to their capability of planning complex action sequences. Sound planning ability is necessary for robust automation in many task domains, but especially in surgical automation. These agents rely on a highly detailed natural language representation of the scene. Thus, to leverage the emergent capabilities of LLM agents for surgical task planning, developing similarly powerful and robust perception algorithms is necessary to derive a detailed scene representation of the environment from visual input. Previous research has focused primarily on enabling LLM-based task planning while adopting simple yet severely limited perception solutions to meet the needs for bench-top experiments, but lacks the critical flexibility to scale to less constrained settings. In this work, we propose an alternate perception approach -- a digital twin (DT) -based machine perception approach that capitalizes on the convincing performance and out-of-the-box generalization of recent vision foundation models. Integrating our DT representation and LLM agent for planning with the dVRK platform, we develop an embodied intelligence system and evaluate its robustness in performing peg transfer and gauze retrieval tasks. Our approach shows strong task performance and generalizability to varied environmental settings. Despite a convincing performance, this work is merely a first step towards the integration of DT representations. Future studies are necessary for the realization of a comprehensive DT framework to improve the interpretability and generalizability of embodied intelligence in surgery.

\keywords{Surgical Intelligence  \and Medical Robotics.}

\end{abstract}
\section{Introduction}
Surgical robots, such as the da Vinci systems, offer enhanced precision, dexterity, control, and visualization, facilitating minimally invasive surgeries that result in fewer complications and faster recovery times. 
With research platforms like the da Vinci Research Kit (dVRK)~\cite{kazanzides2014open} enabling the initial exploration of surgical task automation, emerging language-based automation methods~\cite{moghani2024sufia,brohan2022rt,brohan2023rt} and policy learning methods~\cite{kim2024surgical,fu2024multi} have further accelerated efforts in surgical task automation. Popular tasks to demonstrate surgical automation include peg transfer~\cite{hwang2020efficiently,hwang2022automating,fu2024multi}, suturing~\cite{hari2024stitch,saeidi2022autonomous,kam2023autonomous}, knot tying~\cite{kim2024surgical}, vascular shunt insertion~\cite{dharmarajan2023automating}, and needle picking~\cite{wilcox2022learning,moghani2024sufia} because they offer repeatable testbeds that challenge automation approaches both with respect to task planning and task execution. 
Large language model (LLM)-based automation~\cite{qin2023toolllmfacilitatinglargelanguage,schick2023toolformerlanguagemodelsteach,moghani2024sufia}, in particular, has recently enjoyed particular popularity because LLM agents enable long-horizon planning, potentially in an explainable and interactive way. Capitalizing on the potential benefits of LLM-based automation, however, relies on two key factors: (a) the ability to create detailed scene representations via machine perception, and (b) LLM agent setup to enable task-level planning and control. While previous approaches to LLM-based automation have started to demonstrate promising results, they mostly focus on the latter aspect, how to leverage LLM agents for advanced control planners and policy learning techniques. 
Robust scene representation via machine perception, however, is a critical prerequisite for LLM-based automation. In this work, we present a DT-based approach to LLM-based automation, leveraging robust vision foundation models to extract scene representation from visual input.

\begin{figure*}[ht]
      \centering
      \includegraphics[width=\textwidth]{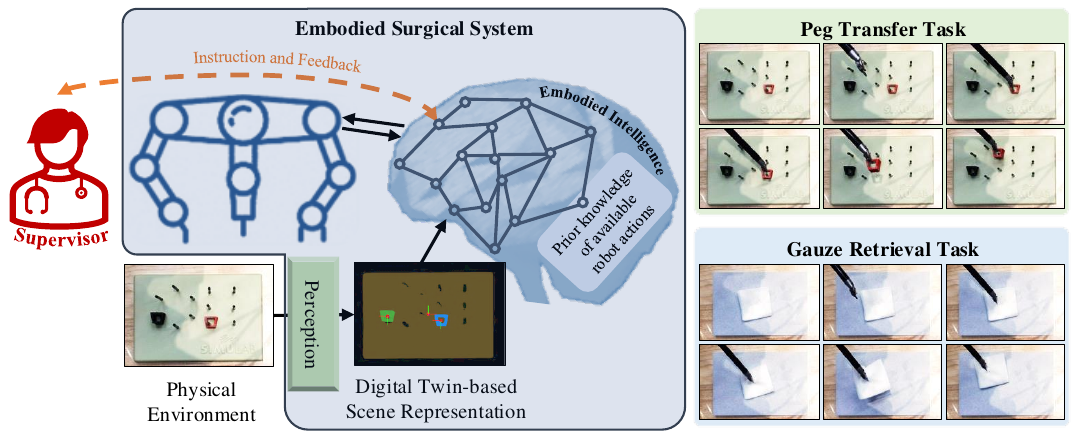}
      \caption{Illustration of the DT-based embodied surgical system. A machine perception module is applied to extract the DT representation from the physical environment. An LLM-enabled embodied intelligence takes commands from a supervisor and makes high-level task plans based on the scene representation, prior knowledge, available actions, and previous actions and feedback. A robotic system receives commands and executes them in the physical world. }
      \label{fig:illustration}
\end{figure*}

Digital twins (DT), computational replicas of the real world (physical twin) created and updated through sensor data analysis, such as machine vision, offer an intermediary layer between the low-level processes (e.g., vision tasks) and the high-level scene analysis and automation tasks. This DT-based paradigm for automation offers a unifying framework for low- and high-level analysis and automation in a more generalizable and interpretable manner~\cite{Ding2024Jul}. To obtain the DT representation, previous works~\cite{shu2023twin,hein2024creating,killeen2024stand,Kleinbeck2024Jul} predominantly relied on external tracking devices and markers to ensure the robustness and accuracy of the system. While advancements in deep learning algorithms for computer vision, such as instance segmentation~\cite{HeGDG17maskrcnn,ChenPWXLSF0SOLL19htc,DingQY021DSC,cheng2021mask2former} and pose estimation~\cite{peng2019pvnet,he20206d,marullo20236d,hein2021towards,li2023tatoo,teufel2024oneslam}, offer an alternate vision-based, marker-less approach to extract the DT representation, these methods lack generalizability and fail when the observed scenario differs from the training data~\cite{ding2022carts,Ding2022RethinkingCR,ding2024segstrongc}. The recent emergence of vision foundation models~\cite{kirillov2023segment,ravi2024sam,wen2024foundationpose,yang2024depth,raiciu2006enabling,xiao2024spatialtracker} offers more generalizable tools for creating DT representations and developing robust machine perception~\cite{shen2024performance,oguine2024generalization}. These advancements can complement powerful LLM-based planners and robot control systems, creating a framework that affords the necessary robustness to accelerate the advancement of surgical task automation.

In this paper, we demonstrate the aforementioned concept by instantiating an embodied surgical system enabled by a basic DT representation. We propose a machine perception module to extract the DT representation robustly. As shown in Fig.~\ref{fig:illustration}, the perception module takes the vision input and extracts the DT representation. The representation is provided to the LLM-enabled embodied intelligence for task planning and further commands the robotic control unit for task execution. We take peg transfer and gauze retrieval as our experimental tasks. We find that our embodied surgical system presents promising automation performance in terms of success rate, exhibiting strong robustness to variations in the experimental environments where rule-based and specifically trained neural network baselines tend to fail.
In summary, our key contributions are:
\begin{itemize}
    \item Proposing a foundation model-based machine perception for extracting DT representation from the physical world.
    \item Proposing an embodied intelligence surgical system, enabled by the DT representation, that presents robust automation performance.

\end{itemize}

\begin{figure*}[!h]
  \centering
  \includegraphics[width=\textwidth]{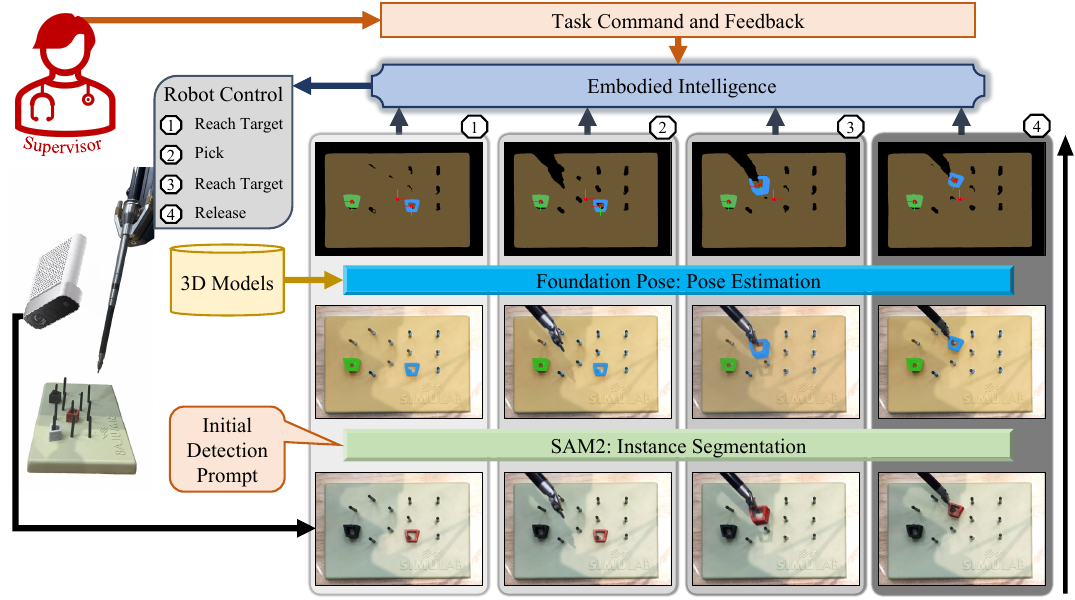}
  \caption{Illustration of the workflow of the proposed embodied surgical system with DT-based machine perception. The captured image is first processed via SAM2~\cite{ravi2024sam} with initial point prompts for the objects of interest. The objects' identification, segmentation, raw image, and corresponding 3D models are processed via the FoundationPose model to predict 6DoF poses. The extracted information forms a DT representation and is further captured by embodied intelligence for task planning.}
  \label{fig:perception}
\end{figure*}

\section{Method}

\subsection{Embodied Surgical System Overview}
Our embodied surgical system incorporates three main components: DT-based machine perception, robotic control system, and embodied intelligence (language-based agent) (Fig.~\ref{fig:illustration}). The DT-based machine perception utilizes RGB-D data extracted from the environment to track objects of interest and generate a basic DT representation of the workspace. The robotic control system applies the da Vinci Research Kit (dVRK)~\cite{kazanzides2014open}, which facilitates the control of the surgical system's Patient Side Manipulator (PSM) to execute the planned action. Taking on the role of embodied intelligence, the language-based agent processes human-level natural language commands and generates corresponding action plans for the robot. These plans are based on the language input, the DT representation, available robot control actions, and real-time feedback.

\subsection{Digital Twin-based Machine Perception}

\paragraph{Digital twin representation}
The DT representation is the quantified information that can be used to construct a DT-based physical environment (physical twin). This representation can encompass identification, geometric, spatial, and physical information like label, shape, pose, and friction. In this work, we apply a basic DT representation using identification, segmentation, 3D models, and 6 DoF poses of the object of interest, which are necessary for automating basic tasks like peg transfer and gauze retrieval.

\paragraph{Perception Workflow}
The DT-based machine perception utilizes the sensory data from an RGB-D sensor to extract the DT representation through a sequential workflow, as shown in Fig.~\ref{fig:perception}. During initialization, 
the Segment Anything Model 2 (SAM2)~\cite{ravi2024sam} is prompted with points that initialize the identification and segmentation of the objects of interest. In the subsequent tracking and update phase, the objects of interest are continuously detected and tracked to update the DT representation in real-time. In this phase, the input RGB-D sensory data is first propagated through the SAM2 model to segment the objects of interest. These segments, along with the 3D model priors of the object and raw image, are then processed by the FoundationPose~\cite{wen2024foundationpose} model to extract the corresponding 6 DoF poses to form the DT representation. 

\subsection{Robotic Control System}
The robotic control system is comprised of the da Vinci Classic surgical system (hardware) and the dVRK \cite{kazanzides2014open} platform (software). The surgical system includes Patient Side Manipulators (PSMs), which are controlled using the dVRK's integrated control system to execute low-level actions (e.g., measured\_cp for forward kinematics, move\_cp for moving in Cartesian space). Before task execution, we first perform a hand-eye calibration using the provided pipelines from dVRK to align the robot's base with the camera coordinates. We use forward kinematics to get the position and orientation in the camera coordinates. 

\subsection{Embodied intelligence}

A language-based agent using GPT4-o is employed to realize embodied intelligence. The system prompt defines the agent's role and provides a set of actions from which the agent can carefully select and sequence to complete a task. Based on the human-level natural language commands, the agent performs step-by-step online planning. At each step, it predicts the next action based on the task command, previous actions, and the supervisor's feedback. Here, the supervisor integrates human feedback into each step to enable closed-loop planning through shared embodiment. 
The set of actions made available to the agent includes perception actions and robot actions, as listed below:

\begin{enumerate}
    \item Get observations: allows the agent to access the extracted DT representation of the environment, such as the identification, segmentation, and pose of objects. Each object is assigned an object ID to aid future planning.
    \item Reach target object: enables the agent to control the surgical system's PSM to reach the pick/place position of an object with a specific object ID. 
    \item Pick target object: allows the agent to close the end-effector (Large Needle Driver) attached to the PSM, to grab/pick the target object. 
    \item Release the object: allows the agent to open the end-effector, releasing a picked object at the current position.
    \item Adjust position: allows the agent to incrementally adjust the robot's position by a fixed offset relative to the camera coordinates based on the specified directions: up, down, left, right, forward, and back. 
    \item Inquiry: allows the agent to interact with the supervisor to get further instructions or clarifications.
\end{enumerate}
After completing a reach/pick action (2, 3), the agent requests feedback from the supervisor to confirm the successful execution of the action. During the reach actions (2), the PSM follows a trajectory based on linearly interpolated waypoints decided from the current and final positions. 

\section{Experiment}
We employ the Azure Kinect RGB-D camera as the vision sensor for machine perception and the dVRK system as the robotic control system~\cite{kazanzides2014open} in our embodied surgical system. We benchmark our system against two baseline methods (Sec.~\ref{baseline_methods}) on the peg transfer task 
, a common laparoscopic training task used for skill training and assessment in surgical training programs. Additionally, the task generalizability of each system is assessed using a gauze retrieval task. 

\begin{figure*}[t]
  \centering
  \includegraphics[width=\textwidth]{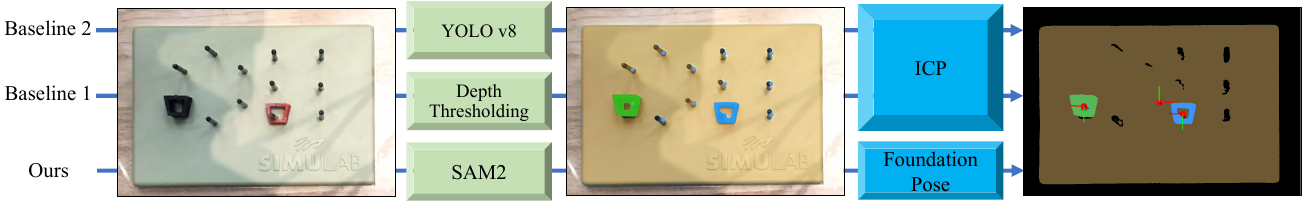}
  \caption{Comparison between our perception module and baseline models.}
  \label{fig:baseline}
\end{figure*}

\subsection{Baseline Methods}
\label{baseline_methods}
We applied two variants as our baseline models. The comparison between our model and baseline models are shown in Figure~\ref{fig:baseline}
\noindent\textbf{Depth thresholding (DTh) + Iterative Closest Point (ICP):} We adopt the depth thresholding + ICP method from Hwang et al.~\cite{hwang2020efficiently,hwang2022automating}. We threshold both the upper bound and the lower bound of the depth to get the target object. The threshold is calculated as $ [min(d_{positive})-\epsilon_{lb}, min(d_{negative})-\epsilon_{ub}] $, where $d_{positive}$ and $d_{negative}$ are the depths for positive and negative prompt points, and ($\epsilon_{lb}$) and ($\epsilon_{ub}$) are the lower and upper bound depth noise tolerance for effective target object-background separation. We initialize translation and rotation to the pose of the back-projected point from the center pixel of the object and identity matrix, respectively. We then apply ICP to refine the final pose of the objects using the projected points and the 3D models.

\noindent\textbf{YOLO + ICP:} We incorporates YOLOv8~\cite{varghese2024yolov8} and ICP. We custom-trained YOLOv8 for instance segmentation on data collected from the ideal experiment setup, with annotation generated by SAM2 and filtered by human annotators. To simulate the initial point prompts provided to the SAM2 model in the other baseline and our method, visible points are added to the images, with distinct colors indicating different objects, for both training and inference. It applies the same pose estimation method using ICP.

\subsection{Peg Transfer in Varied Environments}
We evaluate the robustness of our embodied surgical system, driven by a digital scene representation derived from foundation models, on a peg transfer task. The task involves a pegboard with 12 pegs and some blocks initially placed on the pegs. The robot must pick a specific block and place it on a target peg. One pick-and-place action sequence is considered as one trial. We benchmark our system on both open-loop and closed-loop planning to disentangle the advantages of robust language agents and highlight the effectiveness of our DT-based machine perception. In the open-loop planning framework, the agent plans the actions, and the robotic control system executes them once without any supervisor feedback to verify successful action completion. In the closed-loop planning framework, the agent accepts language feedback from the supervisor. This feedback includes fine-grained position adjustment of the robot end-effector in six directions (up, down, left, right, forward, and back) in the image space,  target re-detection, and re-execution of the action. 
Each position adjustment feedback will adjust the end-effector tool tip position by 3\,mm in the specified direction in camera coordinates. A maximum of 5 position adjustments or redos is allowed for each trial before considering it a failure trial. Both open and closed-loop planning frameworks are evaluated based on the success rate and the failure modes: inaccurate pose (Po), object not detected (De), and planning error (Pl). The closed-loop framework is also evaluated on the number of planning steps. 

\begin{table*}[!h]
\centering
\caption{Experiment Results}
\resizebox{\textwidth}{!}{
    \begin{tabular}{c|c|ccc|cc}
    \toprule
    \multirow{2}{*}{\textbf{Experimental}} & \multirow{3}{*}{\textbf{Method}}    & \multicolumn{3}{c|}{\textbf{Closed-loop planning}}  & \multicolumn{2}{c}{\textbf{Open-loop planning}}
    \\
    \multirow{2}{*}{\textbf{Setup}} &  & \textbf{Success}  & \textbf{Average} & \textbf{Failure Mode} & \textbf{Success} & \textbf{Failure Mode}\\
     &  & \textbf{Rate} & \textbf{Planning Steps} & \textbf{Po, De, Pl} & \textbf{Rate} & \textbf{Po, De, Pl}\\
    \midrule
    \multirow{3}{*}{Ideal Environment}
    & DTh + ICP 
    & ~97\% (~97/100) & 5.59 & ~1,~2,~0
    & ~73\% (73/100) & 25,~2,~0 \\
    & YOLO + ICP
    & ~97\% (~97/100) & 5.64 & ~3,~0,~0
    & ~75\% (75/100) & 25,~0,~0 \\
    & Ours
    & \textbf{100\% (100/100)} & \textbf{5.04} & \textbf{~0,~0,~0}
    & ~\textbf{96\% (96/100)} & \textbf{~4,~0,~0} \\
    \midrule
    \multirow{4}{*}{Black/Red Block}
    & DTh + ICP 
    & 88\% (44/50) &5.80 & 3,~3,~0
    & 46\% (23/50) & 24,~3,~0 \\
    & YOLO + ICP
    & 72\% (36/50) & 5.36 & 8,~6,~0
    & 54\% (27/50) & 17,~6,~0  \\
    & YOLO + FP
    & 90\% (45/50) & 5.04 & 0,~5,~0
    & 86\% (43/50) & ~2,~5,~0  \\
    & Ours
    & \textbf{100\% (50/50)} & \textbf{5.08} & \textbf{~0,~0,~0}
    & $\textbf{96\% (48/50)}$ & \textbf{~2,~0,~0} \\
    \midrule
    \multirow{4}{*}{Tilted Pegboard}
    & DTh + ICP
    & 56\% (28/50) & 6.79 & 11,10,~1
    & ~8\% (~4/50) & 35,10,~1 \\
    & DTh + FP 
    & 78\% (39/50) & 5.23 &  ~1,~10,~0
    & 68\% (34/50) &  ~6,~10,~0 \\
    & YOLO + ICP
    & ~84\% (41/50) & 6.00 & ~7,~2,~0
    & ~36\% (18/50) & 30,~2,~0 \\
    & Ours
    & ~\textbf{96\% (48/50)} & \textbf{5.10} & \textbf{~2,~0,~0}
    & ~\textbf{86\% (43/50)}& \textbf{~7,~0,~0}  \\

    \midrule
    \midrule
    \multirow{4}{*}{Gauze Retrival} & Depth thresholding + ICP & ~84\% (~84/100) & - & 15,~~~0,~1 & - & -  \\
    & YOLO + ICP (Peg transfer data) & ~~0\% (~~0/100) & - & ~0,~100,~0 & - & - \\
    & YOLO + ICP (Gauze data)
    & \textbf{100\% (100/100)} & - & \textbf{~0,~~~0,~0} & - & - \\
    & Ours 
    & \textbf{100\% (100/100)} & - & \textbf{~0,~~~0,~0} & - & - \\
    \bottomrule
    \end{tabular}
}
\label{tab:peg}
\end{table*}

\paragraph{Varied environments}
Our embodied surgical system, in both its open and closed-loop planning configurations, is evaluated against the two baseline models on three varied environments to evaluate the effectiveness of the foundation model-enabled DT representation. 

These environments include:

\begin{itemize}
    \item Ideal environment: The pegboard is positioned at the center of the camera's field of view, with its normal direction perpendicular to the camera plane. We use the grey trapezoid block. 
    \item Changing block color: The block color is changed to black and red.
    \item Changing pegboard orientation: The pegboard is tilted at a fixed angle ($\approx15\degree$) toward the camera plane. 
\end{itemize}
Each method, in each of its planning frameworks, is tested over $100$ trials in the ideal environment and $50$ trials for each varied environment.

\paragraph{Results and Discussion} Table~\ref{tab:peg} quantitatively benchmarks our method against the two baseline methods on varied environments, in both the closed and open-loop planning frameworks. In the closed-loop framework, while all methods achieved a high success rate in the ideal environment, a drop in success rate and an increase in average planning steps are observed for the two baseline models in the other two environments. This indicates that the two baseline methods' machine perceptions are less robust, as indicated by the increase in pose estimation and detection error observed in the respective failure modes. The rise in average planning steps for the baseline models suggests that embodied intelligence is attempting to compensate for the limitations of its machine perception. 

The flexibility and generalizability of our DT-based machine perception, which leverages a foundation model, becomes much more evident when we disentangle (open-loop planning framework) the robustness of embodied intelligence. In the open-loop planning framework, our method outperforms the baseline methods under the ideal environment and varied environments. The limited flexibility of the two baseline methods can be attributed to several factors. The effectiveness of the depth thresholding technique is primarily limited by two main factors: (a) the black color absorbs infrared light, which interferes with depth estimation, and (b) the tilted pegboard makes it harder to threshold the depths between the board and the block. The YOLO model struggled primarily due to the out-of-domain predictions, as the black/red block and tilted pegboard were not included in the training set. As a result, the YOLO model either fails to detect the object or predicts inaccurate segmentation. 

\paragraph{Ablation Studies}
Additionally, we perform ablation studies in varied environments to explore the effectiveness of each component and design choices. We replace ICP with FoundationPose (FP) for YOLO for the black/red blocks environment and for Depth thresholding in the tilted pegboard environment. We use the same setting for varied environments. Results in Table~\ref{tab:peg} show that, although the detection failure cannot be addressed, FoundationPose alleviates the pose estimation error caused by inaccurate segmentation with visual input.

\subsection{Task Generalization: Gauze Retrieval}
To further validate the generalizability of our embodied surgical system, we evaluate its performance on gauze retrieval tasks. This task requires the end-effector to pick up a $5cm\times5cm$ gauze, with each pick-up action considered as a single trial. All methods are evaluated in an open-loop planning framework, based on the success rate in 100 trials. 

From the quantitative results in Table~\ref{tab:peg}, we observe that our method achieves robust performance, demonstrating its zero-shot generalization ability for this task. In contrast, the performance of the baseline method employing the depth thresholding technique declines due to the inseparable depth between the gauze and the background. Similarly, the YOLOv8-based method initially failed to complete the task even once due to the out-of-domain challenges. However, when the YOLOv8 model is further trained on an additional 100 images with gauze annotations, its performance improves to levels comparable with our method.

\section{Conclusion}

With most research on embodied intelligence focusing mainly on advancing language-based agents for robust task planning, we propose an alternate approach focusing on advancing machine perception. We leverage foundation models to extract DT representation to serve as an intermediary layer to complement the LLM-based embodied intelligence and create a flexible, scalable, interpretable, and generalizable surgical embodied system. Our instantiation for the surgical training tasks showcases its potential for robust task automation. 

Besides this, a more comprehensive DT framework allows the generation of massive synthetic data to train high-level scene analysis and automation agents. This, in turn, could potentially enhance the adaptability and generalizability of embodied intelligence in surgery. Challenges remain when apply our idea into soft-tissue manipulation as it requires high-fidelity construction of digital twin representation which requires precise calibration, extensive data integration, and consistent visual fidelity. Thus, future advancements are required in perception, modeling, and simulation in soft-tissue digital twin and more efforts are expected to explore the DT-driven approaches' potential in this regard to advance surgical automation, moving it closer to practical, real-world clinical applications.

\begin{credits}
\subsubsection{\ackname} his research is supported by a collaborative research agreement with the MultiScale Medical Robotics Center at The Chinese University of Hong Kong.
\subsubsection{\discintname}
The authors have no competing interests to declare that are relevant to the content of this article. 
\end{credits}

%
%
%
\bibliographystyle{splncs04}
\bibliography{main}
\end{document}